\documentclass[10pt,twocolumn,letterpaper]{article}

\usepackage{cvpr}              

\usepackage{graphicx}
\usepackage{amsmath}
\usepackage{amssymb}
\usepackage{booktabs}
\usepackage{epsfig}
\usepackage[accsupp]{axessibility}

\usepackage[pagebackref,breaklinks,colorlinks]{hyperref}

\usepackage[capitalize]{cleveref}
\crefname{section}{Sec.}{Secs.}
\Crefname{section}{Section}{Sections}
\Crefname{table}{Table}{Tables}
\crefname{table}{Tab.}{Tabs.}

\def\cvprPaperID{15} 
\def\confName{CVPR}
\def\confYear{2022}

\begin{document}

\title{Density-Guided Label Smoothing for Temporal\\ Localization of Driving Actions }

\author{Tunc Alkanat\textsuperscript{*}, Erkut Akdag\textsuperscript{*,\dag}, Egor Bondarev, Peter H.N. de With\\
VCA Group, Department of Electrical Engineering, Eindhoven University of Technology\\
P.O. Box 513, Eindhoven 5612AZ, The Netherlands\\
\small{*~\textit{equal technical contribution}}\\
\small{\textsuperscript{\dag}~\textit{corresponding author}}\\
{\tt\small \{t.alkanat, e.akdag, e.bondarev, p.h.n.de.with\}@tue.nl}\\
}
\maketitle

\begin{abstract} 
Temporal localization of driving actions plays a crucial role in advanced driver-assistance systems and naturalistic driving studies. However, this is a challenging task due to strict requirements for robustness, reliability and accurate localization. In this work, we focus on improving the overall performance by efficiently utilizing video action recognition networks and adapting these to the problem of action localization. To this end, we first develop a density-guided label smoothing technique based on label probability distributions to facilitate better learning from boundary video-segments that typically include multiple labels. Second, we design a post-processing step to efficiently fuse information from video-segments and multiple camera views into scene-level predictions, which facilitates elimination of false positives. Our methodology yields a competitive performance on the A2 test set of the naturalistic driving action recognition track of the 2022 NVIDIA AI City Challenge with an $F_1$ score of 0.271.
\end{abstract}

\section{Introduction} 
\label{sec:intro}
Efficient, fast, and safe transportation is one of the important foundations of modern society and a major driving factor of the economy. Every day, billions of people rely on various forms of transit, including road transportation. However, frequent use of road transportation also has drawbacks. For instance, the annual road traffic deaths were estimated at 1.25 million in 2013~\cite{world2015global}. Immense research is being conducted on advanced driver-assistance systems~(ADAS) to enhance the safety and comfort of road transportation. 

\begin{figure}[t]
  \centering
  \includegraphics[width=\columnwidth]{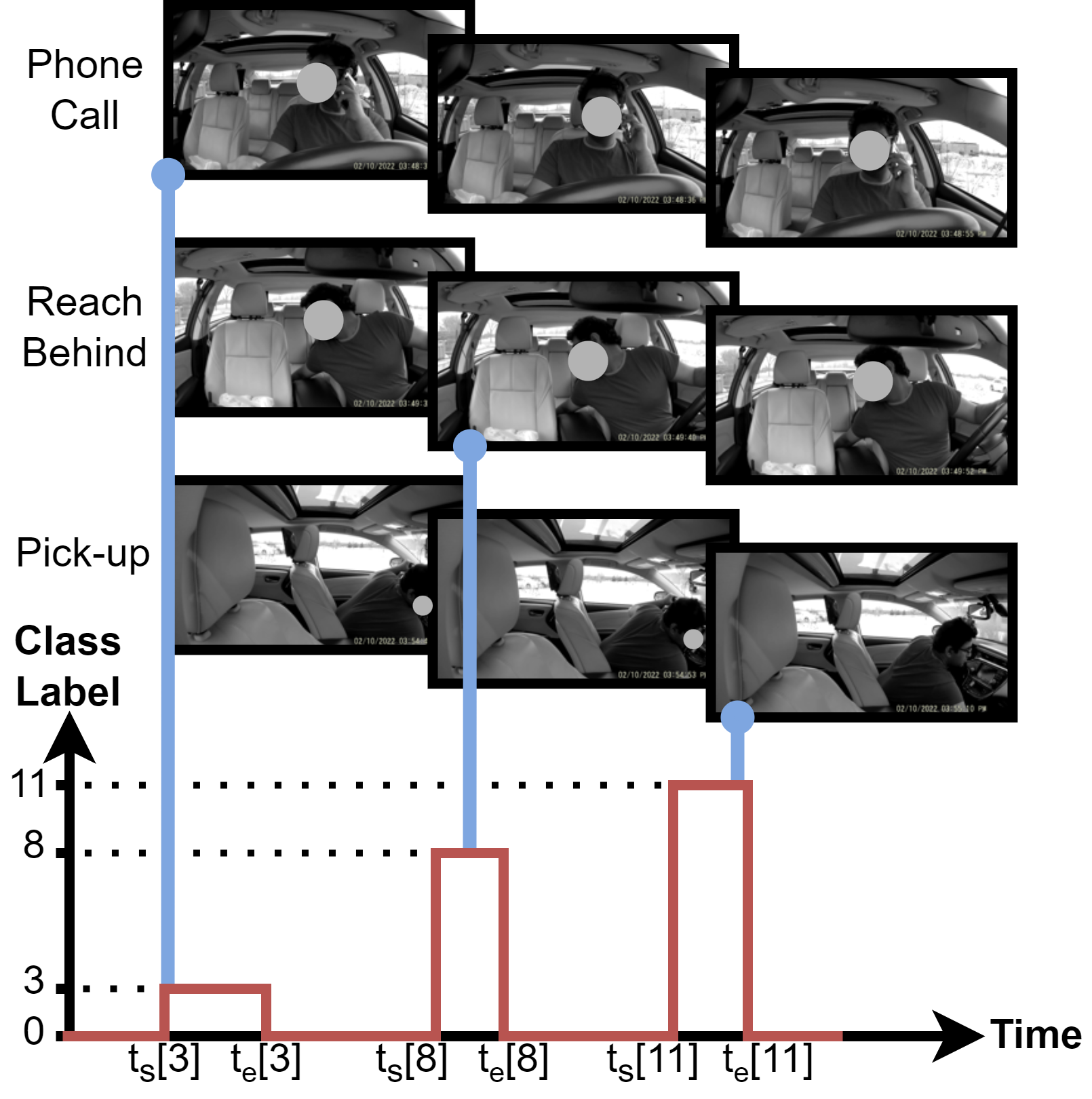} 
  \caption{Diagram of joint task of action classification and localization. From top to bottom: example images of the phone call, reaching behind, and pick up from floor action classes. The graph shows an ideal output, where not only classes but also the start and end times~($t_s[i]$, $t_e[i]$) of actions are visible.
  }
  \vspace{-0.7cm}
  \label{fig:intro}
\end{figure}

The ultimate aim of ADAS is to completely automate road transportation, while ensuring safety by eliminating human factors of road accidents. However, this automation task is difficult due to the complex nature of the problem, variability of environmental conditions, and the strict requirements for high robustness and reliability. In the current state, ADAS requires human supervision to ensure safety while gradually addressing increasingly complex driving tasks. This approach requires continuous attention of the driver, since over-confidence by the driver in such systems may cause deadly accidents. Consequently, the need for real-time monitoring of the driver's alertness arises and novel computer vision methodologies are being researched to satisfy this requirement in an automated fashion.


For a practical implementation of driving action recognition, it is necessary both to correctly classify the specific action of the driver and to temporally localize this action, since the videos in the wild are untrimmed and may include multiple classes at once. This problem is usually referred to as temporal action localization in literature. Many previous works on this subject are inspired by two-stage, proposal-generation and classification methodologies that were initially devised for the problem of object detection~\cite{chao2018rethinking, zeng2019graph}. Alternatively, the straightforward approach of using a temporal sliding window has also been popular in literature~\cite{oneata2014lear,wang2014action}. In the sliding-window approach, action classification is performed for every overlapping segment of the input video, and the resulting class probabilities for each segment are fused in a post-processing step. While this approach provides solid overall performance, the localization accuracy usually suffers from segments that include multiple action classes. To alleviate this shortcoming, in this work, we propose density-guided label smoothing to improve localization performance. The associated loss function considers the distribution of class labels in each training video segment, thereby enabling better control over the performance trade-off of classification and temporal localization sub-tasks. Furthermore, we design a post-processing step for multi-camera fusion and prediction distillation. This design includes the phases of: (1) efficient late-fusion of the predictions across multiple cameras, (2) preservation of sharp temporal changes in class probabilities, and (3) elimination of temporally overlapping detections. To summarize, the main contributions of our work are as follows.
\begin{itemize}
    \item A smoothed multi-label training loss which facilitates better learning from temporal boundary segments. The improved performance comes without any additional computational overhead during inference.
    \item An efficient post-processing step that eliminates false positives and improves the overall performance.
    \item An evaluation of the proposed method on the A2 test set of the naturalistic driving action recognition track of the 2022 NVIDIA AI City Challenge. Our method shows promising results with an $\text{F}_1$ score of 0.271.
\end{itemize}


\section{Related Work} 
\label{sec:rw}
This section presents a literature overview of video feature extraction, temporal action localization, and driving action recognition datasets.

\vspace{0.2cm}
\textbf{Video feature extraction.} Extracting features from given videos is a crucial first step to many advanced scene understanding tasks, such as action recognition and video anomaly detection. As such, this problem has been extensively studied. Early approaches to the problem focus on handcrafted motion features, such as Hidden Markov Models~\cite{hospedales2009markov,kratz2009anomaly}, sparse coding~\cite{zhao2011online,mo2013adaptive}, histogram of oriented gradients (HOGs)~\cite{chaudhry2009histograms, colque2016histograms, li2013anomaly}, and appearance features~\cite{cong2011sparse,li2013anomaly} Besides handcrafted feature extraction, deep learning-based methods have been recently proposed for video action recognition tasks. Especially convolutional neural network-based~(CNN) architectures~\cite{chen2021deep} have provided solid performance. Various 2D-CNN methods~\cite{wang2016temporal,lin1811temporal,fan2019more}, 3D-CNN methods, such as C3D~\cite{tran2015learning}, I3D~\cite{carreira2017quo} and ResNet3D~\cite{hara2018can}, and combinations of both~\cite{luo2019grouped, sudhakaran2020gate} are used for spatio-temporal feature extraction. Similarly, the SlowFast~\cite{feichtenhofer2019slowfast} network processes videos at both low and high frame rates to simultaneously capture short-term and long-term temporal information by employing two pathways.

\textbf{Temporal action localization.} Revealing the temporal locations of important events in untrimmed videos is a challenging  task. Early approaches~\cite{karaman2014fast,wang2014action,yuan2016temporal} apply temporal sliding windows as an exhaustive search solution to the localization, which is followed by a support vector machine~(SVM) to classify actions within each window position. Besides sliding-window approaches, object detection-inspired~\cite{ren2015faster} proposal generation methodologies have been employed. For instance, in \cite{chao2018rethinking}, authors use dilated convolutions to encode temporal context and perform multi-stream feature fusion to improve action localization. In~\cite{zeng2019graph}, Zeng~\etal concentrate on capturing the context information and characterize the correlations between distinct actions by implementing Graph Convolutional Networks~(GCNs).

\begin{figure*}[ht]
  \centering
  \includegraphics[width=\linewidth]{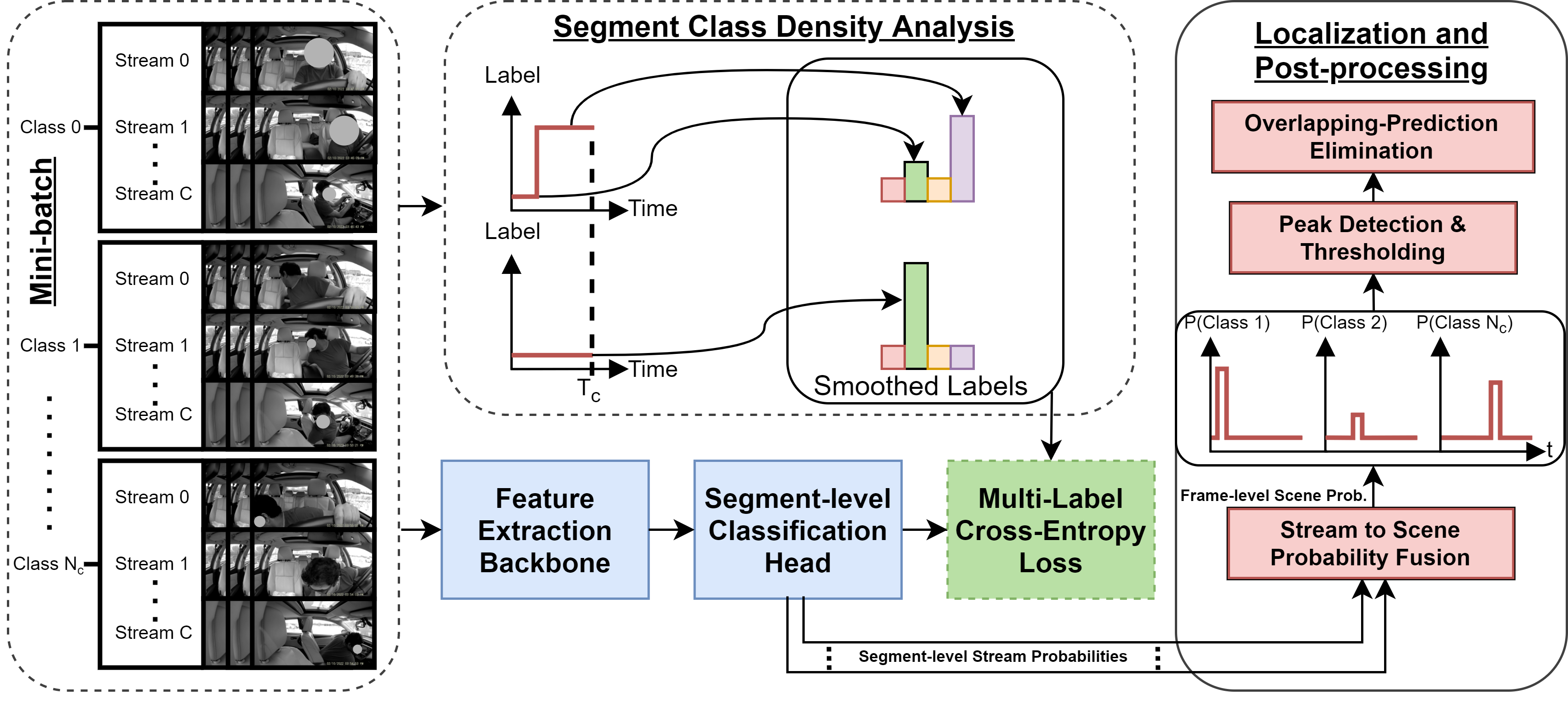} 
  \caption{Overview of the methodology. Left: multi-class, multi-view structured mini-batching ensures a training without bias. Every stream is a camera view. Middle: the network is trained using cross-entropy loss, where the target probabilities are computed individually for every segment according to the class density within that segment. Right: localization and post-processing concatenates the class probabilities of streams into the scene class probabilities by fusing streams. Then, the 1-D probability signal is analyzed for peaks for every class in a scene. Significant peaks are considered predictions and are further refined by eliminating the overlapping predictions. Note that the modules shown with a dashed border are used for training only.}
  \label{fig:framework}
\end{figure*}

\textbf{Driving action recognition datasets.} Quality of datasets is of utmost importance, especially for the data-hungry deep learning-based approaches. Some datasets that comprise of in-vehicle camera footage have been made publicly available for research on naturalistic driving studies. For instance, the Driver Anomaly Detection~(DAD) dataset~\cite{kopuklu2021driver} consists of 31~subjects that perform various activities in a real car. The dataset is recorded synchronously from the front and top view. This dataset includes approximately~538 and 145~minutes of video recording for normal driving and anomalous driving, respectively. Another important dataset in literature is called Drive\&Act~\cite{martin2019drive}. This dataset includes twelve distraction-related actions. The data include 9.6~million frames captured by six cameras and three imaging modalities. The distracted driver dataset~(DAD) is published in another study~\cite{eraqi2019driver} which involves 44~drivers. The dataset includes 14478~frames distributed over the following classes: safe driving, phone right/left, text right/left, adjusting the radio, drinking, hair/makeup, reaching behind, and talking to the passenger.

From the related work study, we adopt the SlowFast feature extraction along with the sliding-window approach and enhance it with a loss function that improves temporal localization.

\section{Methodology} 
\label{sec:method}
Our methodology is depicted in \Cref{fig:framework}. It consists of three subsequent parts: (1) feature extraction, (2) segment-level classification, and (3) localization and post-processing.

\subsection{Feature Extraction}
Our method relies on the temporal sliding-window approach and its associated video segment features to detect and localize distracted driving behaviors. We adopt this approach in order to benefit from the mature video action recognition literature, to exploit pre-training on large-scale datasets in the field, and to be able to capture long-term, complex motion cues. Without loss of generality, we use the SlowFast~\cite{feichtenhofer2019slowfast} video recognition network as our backbone for feature extraction. The multi-branch architecture of the SlowFast network simultaneously captures both long-term and short-term relations, by applying temporal sub-sampling to the given video segments to extract one low~(fast pathway) and one high~(slow pathway) frame rate input sequence, as in the original paper. The resulting fast pathway focuses on long-term actions and vice versa. We choose this architecture for feature extraction, since the given distracted driving classes are of varying continuity. For instance, the action classes of \textit{eating} and \textit{drinking} are usually of intermittent nature, while the \textit{phone call~(right/left)} and \textit{reaching behind} actions show a temporal continuity. Multiple temporal resolutions considered by the SlowFast network enable robust feature extraction for both cases.

The SlowFast backbone receives all video segments from the given training videos. Formally, assume that $V_t^i\in\mathbb{R}^{H\times{}W\times{}T_c}$ denotes a video segment that consists of $T_c$ frames at the time interval $[t, t+T_c-1]$ of the $i^{th}$ video with horizontal and vertical resolutions of $H$ and $W$, respectively. We remove the classification head of the SlowFast architecture to extract the high-level and descriptive features from both the slow and fast pathways. We then combine the branch features by averaging. The resulting feature vector, $f_t^i = S(V_t^i)\in\mathbb{R}^{N_f}$, is used further as the segment features, where $S$ is the functional form of the SlowFast backbone and $N_f$ is the resulting feature size.

\begin{figure}[ht!]
  \centering
  \includegraphics[width=\columnwidth]{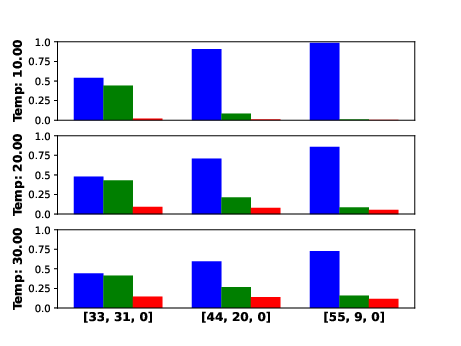} 
  \caption{Bar chart depicting the effects of the temperature parameter $\beta$ and $f(x, y_s)$ on the computed density-guided smooth labels, shown as the bar height. For this example, $N_c=3$ and $T_c=64$ and $\beta$ is set to 10, 20, and 30, from top to bottom. The number of frames for each class is given along the horizontal axis.
  }
  \vspace{-0.5cm}
  \label{fig:temp}
\end{figure}

\subsection{Segment-level Classification}
SlowFast network is deployed with weights pre-trained on the Kinetics-400 video action recognition dataset~\cite{kay2017kinetics}. This large-scale dataset includes 400 classes of various human actions and, thus, ensures that the network focuses on important spatio-temporal aspects of the data related specifically to the driver in the scene. We adapt the pre-trained SlowFast network to the problem at hand by adding fully-connected layers to facilitate the target-specific learning. Explicitly, the segment-level features extracted from the SlowFast backbone, $f_t^i$, are input to two fully-connected layers that generate probabilities for each class, $p_t^i(y)$, where $y\in[1, N_c]$ denotes the class labels. The activation functions of the first and the second fully-connected layer are set to sigmoid and softmax, respectively. 

\textit{Data Imbalance.} Training data imbalance usually leads to performance degradation in classification tasks. To address this problem, we train the model with a sampler that ensures a mini-batch to have equal amount of samples per class and per camera view. Since we use frame-level labels for training and the video segments may contain more than one class, the label with the maximum number of frames in a given segment is regarded as the label for the entire video segment for the mini-batch sampling. On the other hand, ensuring a balanced sampling of data from each camera view improves stability of training and results in a final model that is unbiased against input streams.

\textit{Boundary Segments.} One important pitfall of the temporal sliding-window approach lies in how multi-label video segments are utilized during training. Such segments present a challenge to the network if the training is posed as single-label classification. In such a case, the common approach is to either use the most commonly observed frame-level label as the segment-level label during training, or to discard heterogeneous segments altogether. However, we conjecture that the overall performance of the network can be improved if the information contained within the boundary segments are better utilized. For this, we develop the density-guided label smoothing technique.

\subsubsection{Density-guided Label Smoothing}
Label smoothing is a commonly used regularization technique in machine learning. The aim is to prevent over-confidence of the model in its predictions and thus to improve the overall generalization. In the work~\cite{szegedy2016rethinking}, the mathematical formulation of the label smoothing is given as
\begin{equation}
    q^{\prime}(k|x) = (1-\epsilon)\delta_{k, y} + \epsilon u(k),
\end{equation}
where $y$ is the ground-truth label for a given training sample $x$, $q^{\prime}(k|x)$ is the smoothed ground-truth distribution over labels, $\delta_{k, y}$ is the Dirac delta function which equals unity for $k=y$ and zero otherwise, $u(k)$ is a distribution over labels, and $\epsilon$ controls the amount of smoothing. Usually, the $u(k)$ distribution is selected as a uniform distribution with a probability $1/N_c$.

The label smoothing improves generalization of the model and increases tolerance to label noise. Overall, this approach yields faster convergence and better performance for many classification tasks. Here, starting from this idea, we introduce density-guided label smoothing to facilitate better learning from the boundary segments and achieve a superior overall localization performance. Our technique analyzes the distribution of frame-level labels in each considered segment to compute the smoothed labels. The number of frame-level labels in a segment can be represented by
\begin{equation}
    f(x, y_s) = \sum_{m=1}^{T_c}\delta_{y_{x_m}, y_s},
\end{equation}
where $y_{x_m}$ is the frame-level label of m\textsuperscript{th} frame of the sample video segment, $x$, and $f(x, y_s)$ denotes the total count of frames with the label $y_s$ in $x$. Then, we derive the density-guided smooth labels for a given sample $x$ as
\begin{equation}
    q^{\prime\prime}(k|x) = \frac{e^{\frac{1}{\beta}f(x,k)}}{\displaystyle\sum_{j=1}^{N_c}e^{\frac{1}{\beta} f(x, j)}}.
\end{equation}
In the above equation, we apply the generalized form of the softmax function with the temperature parameter $\beta$. The generalized softmax has several important properties beneficial for the task: (1) its values are guaranteed to be within [0; 1], regardless of the values of $f(x, y_s)$, (2) it sums up to unity when summed for all values of $k$, and (3) via the parameter $\beta$, it enables the control over smooth vs. sharp transitions of the label change.

A bar chart depicting the effects of the temperature parameter and $f(x, y_s)$ on target labels is presented in \Cref{fig:temp}. As can be observed from the figure, the temperature parameter is effective in controlling the target values to get the desired balance between multi-labels and can function as regular label smoothing for the classes that are absent in a given segment.

\textit{Loss Function.} We train our network with the cross-entropy loss for classification. Including the density-guided label smoothing, the complete loss function is given as
\begin{equation}
    \mathcal{L}(x) = -\sum_{k}p_t^i(k)q^{\prime\prime}(k|x),
\end{equation}
where $(i, t)\in B$ defines a unique video segment in a mini-batch~$B$.  After training, we supply test video segments to our network, extract segment class probabilities, $p_t^i(y)$, which are then further subject to the subsequent post-processing.

\subsection{Post-processing}
\label{ssec:pp}
Unlike most object detection-inspired models, the sliding-window-based approaches require a post-processing step to construct the final predictions. Our post-processing pipeline consists of three consecutive parts, (1) fusion of the stream probabilities, (2) peak detection and thresholding, and (3) elimination of overlapping predictions.   

\textit{1. Late-fusion of the Stream Probabilities.} Our methodology can be applied to both multi-camera and single-camera scenarios. However, while the single-camera case is straightforward, the multi-camera setting requires the fusion of individual stream probabilities into frame-level \textit{scene probabilities}. In other words, the class probabilities of all synchronized camera views should be combined~(fused) to extract class probabilities of the multi-camera scene under consideration. Note that, every scene in the dataset has multiple streams~(camera views). For this fusion, we average all the segment-level class probabilities associated with a video segment that contains the considered frame. This step computes the frame-level probabilities by combining the segment-level probabilities which are derived by the network. Here, we extend the previous works and use the output of the softmax activation function instead of raw class scores. Consequently, each of the resulting normalized class probabilities exhibits the inter-class correlation. As a result, the final likelihood value assigned to a combination of a class label and a specific scene frame includes the information regarding the likelihood of other class probabilities for that frame.

\textit{2. Peak Detection and Thresholding.} Raw predictions of temporally localized and classified actions can be obtained by detecting consistent peaks in the frame-level scene probabilities for each class and scene separately. To detect consistent peaks in this 1-D signal, first, we apply a temporal median filtering to eliminate noise, which preserves sharp edges that are helpful in accurate localization of actions. Second, we determine the highest peak of the filtered signal of probabilities. If the observed peak is higher than a pre-defined probability threshold ($\tau$), we consider that peak as a detected relevant action. Finally, we determine the fastest positive and negative changes that precede and succeed the highest peak. The resulting time instances at which the fastest changes in probability occur are then considered as the start and end times of the detected action. \Cref{fig:results} visualizes the step of peak detection. 

\textit{3. Elimination of the Overlapping Predictions~(EOP).} Due to the similar appearance of some actions, it is possible that more than one action are assigned high probabilities for the same frame in individual scenes. However, since the target dataset includes only one action for a time instance in all scenes, overlapping predictions result in an increased false-positive rate and consequently decrease the overall precision of the methodology. To overcome, we first determine the intersection-over-union~(IoU) for all possible pairs of the same-scene predictions. Then, among the full set of predictions with an IoU overlap greater than a pre-defined threshold ($o_{max}$), we only retain the prediction with the highest peak in the final output. 

\section{Experiments} 
\label{sec:exp}
\subsection{Track3 Dataset}
The 2022 NVIDIA AI City Challenge presents naturalistic driving data~\cite{dset} containing 90~video clips (14~hours in total) recorded from different angles by three synchronized cameras mounted in a car. Each video is about 10~minutes long with a resolution of 1920$\times$1080~pixels. The dataset consists of 15~drivers with and without appearance blocks (e.g. sunglasses, hat) performing 18~different tasks (e.g. phone call, eating, and reaching back) once, in random order throughout a video clip. The whole dataset is split into three: A1, A2, and B, each containing five drivers. While the A1~dataset is provided with the ground-truth labels of the start time, end time, and the type of distracting behaviors, the A2~dataset is provided without labels. The main target of the challenge is to classify the distracted behavior activities executed by the driver in a given time frame on the A2~testing dataset, by employing the A1~dataset for training. Organizers reserve the B~dataset for later testing to consider the final awards for the top performers. 

\subsection{Implementation Details}

\textit{1. Feature Extraction.}
We adopt the SlowFast model, pre-trained on the Kinetics-400 video action recognition dataset, to our task of obtaining the feature vectors. As a data-processing step, we re-size and crop each video frame to 256$\times$256~pixels and feed the SlowFast model accordingly, by repeating this process for each 64-frame segment of the video clip with a temporal stride~1. The size of output feature vector, $N_f$, is 2,304 for each segment.

\textit{2. Segment-level Classification.}
The extracted feature vectors $f_t^i$ are given as an input to the first layer of two fully-connected layers, where the sigmoid activation function is applied to a hidden layer size of 64. The last layer generates probabilities for each action class by utilizing the softmax activation function. We adopt the Adam optimizer with initial learning rate 5$\times{10}^{-5}$ and weight decay 5$\times{10}^{-4}$. As the loss function, the cross-entropy loss with density-guided label smoothing is used in our proposed model. The temperature parameter, $\beta$, is set to 5. We employ the GTX-1080 GPU for the training process and set the batch size to~270, which handles 15~training samples for each action class within three camera views (e.g. rear, dash, and right).

\textit{3. Localization and Post-processing.}
We apply the temporal median filtering with a filter size of 301~frames. Minimum height and width for the peak detection are set empirically to 0.05, and 200~frames, respectively.

\subsection{Evaluation Metrics}
Evaluation for 2022 NVIDIA AI City Challenge Track3 is based on the model action identification performance, measured by the $F_1$ score. For computing the $F_1$ score, a true positive (TP) action identification is considered when an action is correctly identified as starting within one second of the start time and ending within one second of the end time of the action. A false positive (FP) is an identified action that is not a TP action. Finally, a false negative (FN) action is a ground-truth action that is not correctly identified. The $F_1$ score is the harmonic mean of recall $R_c$ and precision $P_r$, specified by
\begin{equation}
  F_1 = 2 \times \frac{P_r \times R_c}{P_r+R_c}.
  \label{eq:f1score}
\end{equation}
The precision is calculated as the number of true positives divided by the total number of true positives and false positives, which is given by:
\begin{equation}
  P_r = \frac{\text{TP}}{\text{TP}+\text{FP}}.
  \label{eq:precision}
\end{equation}
The recall is calculated as the number of true positives divided by the total number of true positives and false negatives, which is given by:
\begin{equation}
  R_c = \frac{\text{TP}}{\text{TP}+\text{FN}}.
  \label{eq:recall}
\end{equation}
Although the final results of the challenge are announced based on the $F_1$ score, the evaluation system also presents recall and precision metrics. 

\subsection{Experimental Results}
\begin{figure*}
  \centering
  \includegraphics[width=0.96\linewidth]{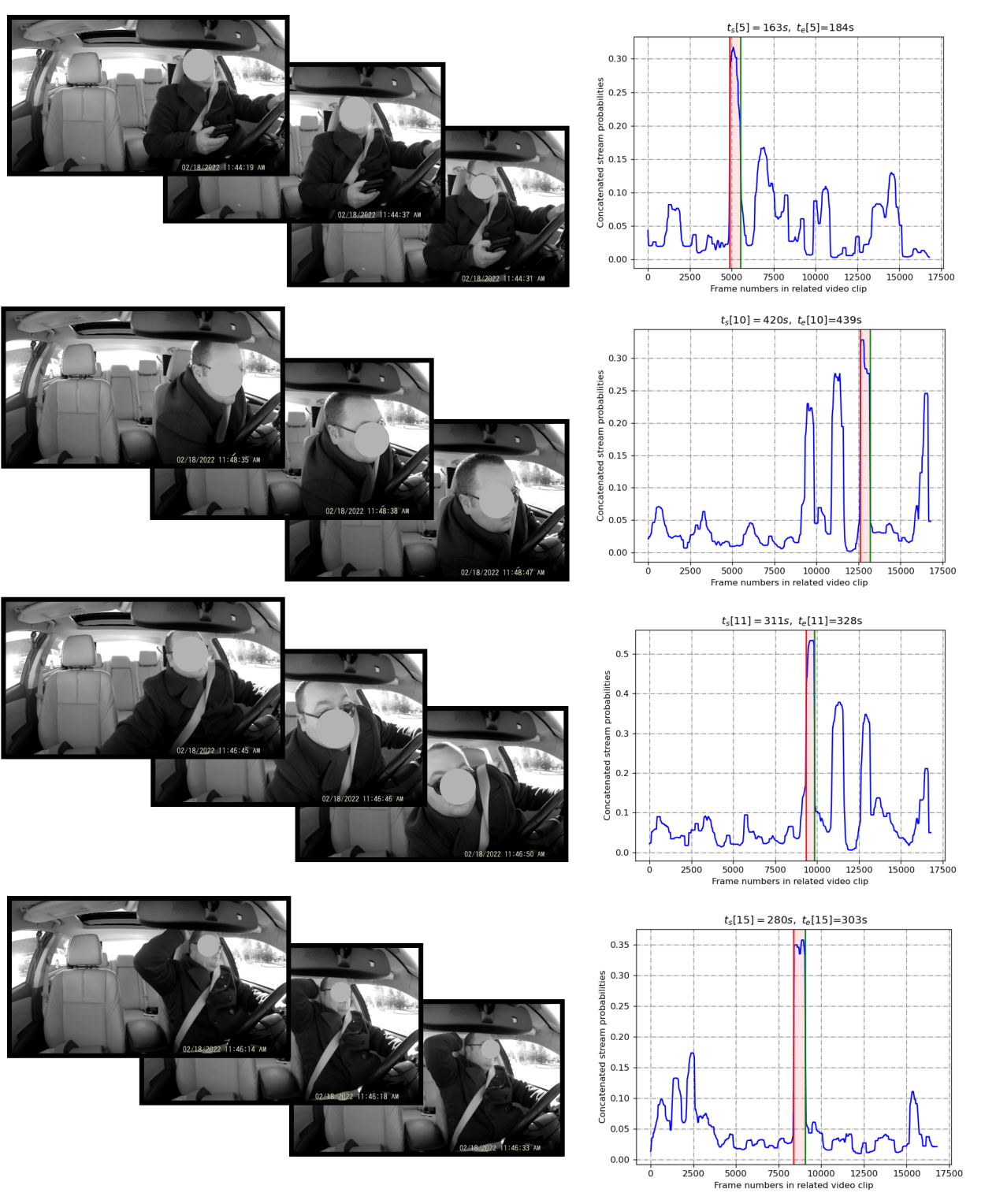} 
  \caption{Experimental results of our methodology. From top left to bottom: example images of texting by the right hand, pick up from the floor on the driver side, pick up from the floor from the passenger side, hand on head action classes. From top right to bottom: graphs show the predicted start and end times~($t_s(i)$, $t_e(i)$) for each class of the video clip. After obtaining the concatenated stream probabilities for each frame within the video clip, we apply the localization and post-processing steps to locate each class's start and end time.}
  \label{fig:results}
\end{figure*}

We evaluate the proposed methodology on the 2022 NVIDIA AI City Challenge Track3 A2 test dataset. Some of the experimental results are summarized in \cref{fig:results}. In this figure, example images are shown from the A2 test dataset, from top left to the bottom, of texting~(right), pick-up from the floor~(driver), pick-up from the floor~(passenger), and hand-on-head action classes. The example pictures, including the sunglasses appearance block, are presented only for the rear-view camera, since all camera views are synchronized. From top right to the bottom, we display graphs including the start and end time of each class after localization and post-processing steps. The vertical axis of the each graph indicates the computed concatenated stream probabilities of corresponding frames, while the horizontal axis shows the frame numbers within the video clip per class. As shown in \cref{fig:results}, the proposed methodology can predict the first and last action classes' start and end times with confidence. Although other peaks appear in the second and third graphs, the proposed algorithm still delivers the result correctly with the help of the stages on peak detection \& thresholding and overlapping-prediction elimination. Overall, visual inspection of the corresponding video data verifies that most of the predictions of our model were accurately localized and classified.

We achieve 0.2710~$F_1$ score and rank the 9\textsuperscript{th} place among 27~teams in Track3, as can be seen from \cref{tab:leaderboard}. In addition to the leaderboard results, we were able to test only a subset of our contributions on the evaluation system. For instance, \cref{tab:results} shows the improvement of performing the overlapping-prediction elimination step within the localization and post-processing stage. 

\begin{table}[h]
  \centering
   \caption{Improvement of overlapping-prediction elimination step on the general leaderboard.}
  \begin{tabular}{@{}lccc@{}}
\toprule
Method   & $F_1$ score & Precision & Recall \\ \midrule
Baseline & 0.2636      & 0.2706    & 0.2570 \\
BL + EOP & 0.2710      & 0.3206    & 0.2346 \\ \bottomrule
\end{tabular}
  \label{tab:results}
\end{table}

\begin{table}[ht]
  \centering
  \caption{Final results based on $F_1$ score on 2022 NVIDIA AI City Challenge Track3 from top-15 on the public leaderboard.}
  \begin{tabular}{@{}lclc@{}lc@{}lc@{}}
    \toprule
    Rank & Team ID & Team Name & $F_1$ \\ 
    \midrule
    1 &72 & VTCC-UTVM & 0.3492\\
    2 &43 & Stargazer & 0.3295 \\
    3 &97 & CybercoreAI & 0.3248 \\
    4 &15 & OPPilot & 0.3154 \\
    5 &78 & SIS Lab & 0.2921 \\
    6 &16 & BUPT-MCPRL2 & 0.2905 \\
    7 &106 & Winter is Coming & 0.2902 \\
    8 &124 & HSNB & 0.2849 \\
    \textbf{9} &\textbf{54} & \textbf{VCA} & \textbf{0.2710} \\
    10 &95 & Tahakom & 0.2706 \\
    11 &1 & SCU\_Anastasiu & 0.2558 \\
    12 &148 & union & 0.2301 \\
    13 &76 & Starwar & 0.2160 \\
    14 &85 & Aespa winter & 0.1440 \\
    15 &69 & SEEE-HUST & 0.1348 \\
    \bottomrule
  \end{tabular}
  \label{tab:leaderboard}
\end{table}

\subsection{Discussion}
Considering the correctly identified distracting activity as starting within one second of the start time and ending within one second of the end time of the activity, which is defined as part of the evaluation criteria, makes this detection task significantly difficult. One of the related challenges is that accurately defining the ground-truth labels of the start and end times can be subjective for some activities. For instance, in a drinking activity, some may annotate the start time as touching the bottle, while others may annotate the moment the liquid enters the mouth, or even the moment of reaching out to the bottle. This ambiguity of timing makes the definition of the action label uncertain possibly leading to false model prediction because those examples may differ by two to three seconds. Another challenge is that the range of possible distracted behaviour types is broader than the given set of classes, e.g. listening to instructions, organizing the objects on the passenger seat, and taking off the mask. As a consequence, these actions can be easily mismatched with the defined classes of the dataset. 

As a possible improvement for alleviating some ambiguity, the audio modality of three camera view recordings may be exploited as well. Audible information can be introduced as additional data without large extra cost. The model based on the proposed methodology can then learn better by jointly analyzing the audio and video datasets.

\section{Conclusion} 
\label{sec:conc}
In this paper, we proposed a methodology that localizes and classifies distracted driver behavior from in-vehicle video data. This approach relies on the fusion of CNN-based video segment features. To enable better learning, we have developed a novel technique called the density-guided label smoothing, which improves the localization performance. Furthermore, we have designed a post-processing step, which efficiently combines data from multiple cameras and eliminates overlapping predictions by retaining only the most confident prediction. This filtering of the predictions improves the precision of the methodology by eliminating false positives. Performance evaluation of our methodology on the A2 test set of the naturalistic driving action recognition track of the 2022 NVIDIA AI City Challenge yields competitive results.  

\section*{{Acknowledgements}}
This work is supported by the European ITEA project SMART on intelligent traffic flow systems and the NWO Efficient Deep Learning~(EDL) RMR project.

{\small
\bibliographystyle{unsrt}

}

\end{document}